\documentclass{article}
\usepackage{spconf,amsmath,graphicx,amsfonts}

\newcommand{\etal}{\textit{et al.}}

\usepackage{multirow}
\usepackage{verbatim}
\usepackage{arydshln}
\usepackage{caption}
\usepackage[labelformat=simple]{subcaption}

\usepackage{stfloats,enumitem}
\usepackage[dvipsnames]{xcolor}
\usepackage{soul}
\usepackage{hyperref}

\title{Open-Set Recognition with Gradient-Based Representations}
\name{Jinsol Lee and Ghassan AlRegib}
\address{
School of Electrical and Computer Engineering,\\ Georgia Institute of Technology, Atlanta, GA, 30332-0250\\ \{jinsol.lee, alregib\}@gatech.edu}
\begin{document}
\twocolumn[{%
\vspace{30mm}
{ \large
\begin{itemize}[leftmargin=2.5cm, align=parleft, labelsep=2cm, itemsep=4ex,]

\item[\textbf{Citation}]{J. Lee and G. AlRegib, “Open-Set Recognition with Gradient-Based Representations,” in \textit{IEEE International Conference on Image Processing (ICIP)}, Anchorage, Alaska, USA, 2021.}

\item[\textbf{Review}]{Date of Publication: September 19, 2021}


\item[\textbf{Bib}]  {@inproceedings\{lee2021osr,\\
    title=\{Open-Set Recognition with Gradient-Based Representations\},\\
    author=\{Lee, Jinsol and AlRegib, Ghassan\},\\
    booktitle=\{IEEE International Conference on Image Processing (ICIP)\},\\
    year=\{2021\}\}}

\item[\textbf{Copyright}]{\textcopyright 2021 IEEE. Personal use of this material is permitted. Permission from IEEE must be obtained for all other uses, in any current or future media, including reprinting/republishing this material for advertising or promotional purposes, creating new collective works, for resale or redistribution to servers or lists, or reuse of any copyrighted component of this work in other works.}

\item[\textbf{Contact}]{
\{jinsol.lee, alregib\}@gatech.edu\\
\url{https://ghassanalregib.info/}\\}
\end{itemize}
}}]
\newpage
\clearpage

%
\maketitle
\begin{abstract}
Neural networks for image classification tasks assume that any given image during inference belongs to one of the training classes. This closed-set assumption is challenged in real-world applications where models may encounter inputs of unknown classes. Open-set recognition aims to solve this problem by rejecting unknown classes while classifying known classes correctly. In this paper, we propose to utilize gradient-based representations obtained from a known classifier to train an unknown detector with instances of known classes only. Gradients correspond to the amount of model updates required to properly represent a given sample, which we exploit to understand the model's capability to characterize inputs with its learned features. Our approach can be utilized with any classifier trained in a supervised manner on known classes without the need to model the distribution of unknown samples explicitly. We show that our gradient-based approach outperforms state-of-the-art methods by up to 11.6\% in open-set classification.
\end{abstract}


\section{Introduction}

Despite the significant advancement in many applications of deep neural networks, they are known to be prone to failure when deployed in real-world environments as they often encounter data that diverges from training conditions~\cite{temel2018cureor, temel2019multifarious}. They rely heavily on the implicit closed-world assumption that any given input during inference belongs to one or more of the classes in training data. Limited to the \textit{knowns} defined by training set, neural networks classify any input images to be among the known classes, even if given inputs are significantly different from training data. In addition, neural networks tend to make overconfident predictions even for the unfamiliar inputs~\cite{goodfellow2014adversarial,guo2017calibration}, making it more challenging to distinguish the \textit{unknowns} from the \textit{knowns}. These types of behaviors of neural networks can have serious consequences when utilized in safety-critical applications, such as autonomous vehicles and medical diagnostics. 

\begin{figure}[t]
    \centering
    \includegraphics[width=0.98\linewidth]{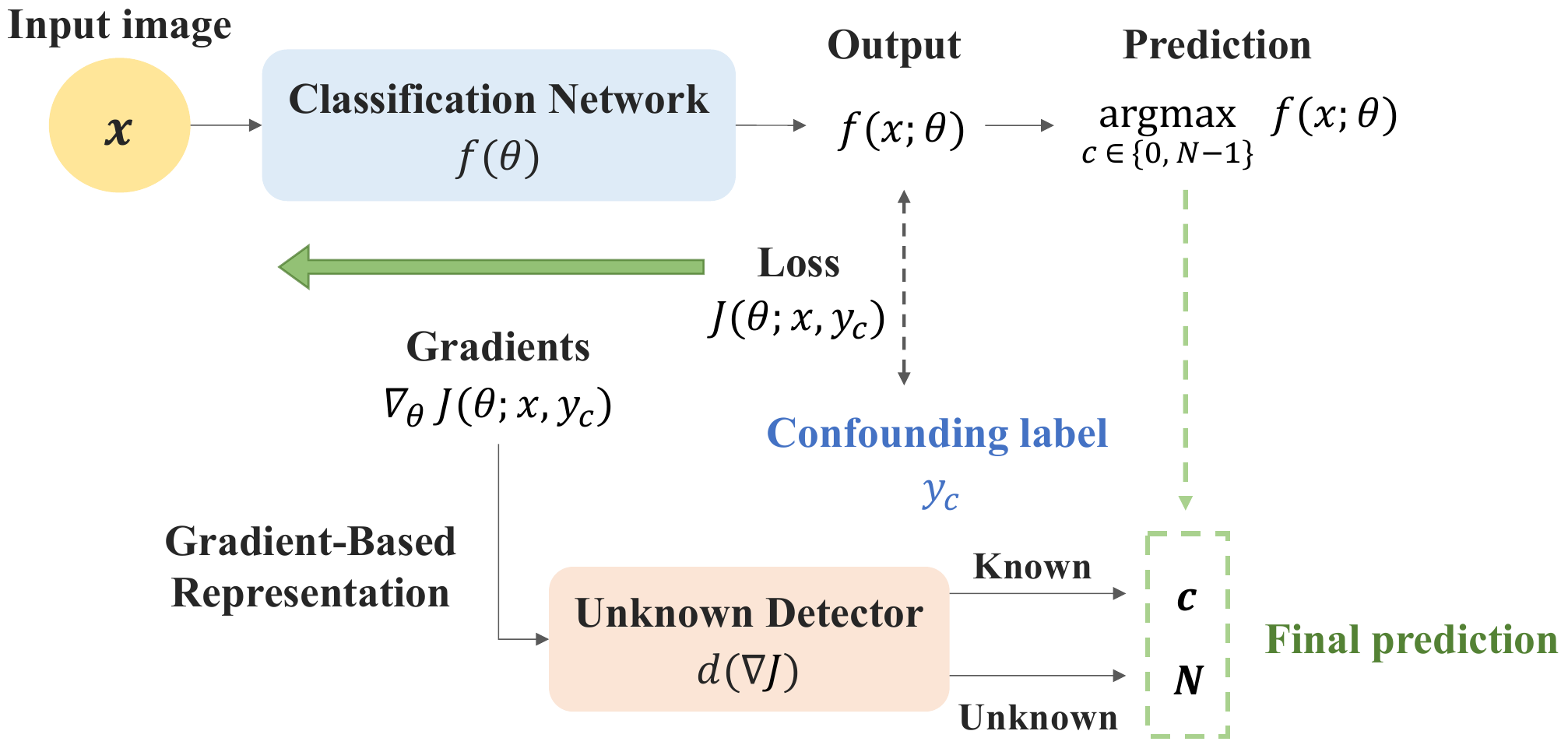}
    \caption{Open-set recognition framework with gradient-based representations generated with confounding labels.}
    \label{fig:framework}
\end{figure}

Open-set recognition tackles this problem by removing the closed-world assumption. Instead, an open-set classifier assumes that testing samples may come from any class, even unknown during the model training. Most approaches in the literature can be divided into two categories: discriminative models and generative models. Discriminative modeling approaches~\cite{scheirer2012toward_osr, bendale2016towards_osdn, yoshihashi2019crosr, oza2019c2ae, sun2020cgdl} aim to learn the distribution of known samples to distinguish between the known classes for classification as well as between the known and unknown classes for unknown detection. Generative modeling approaches~\cite{ge2017generative_openmax, neal2018counterfactual} seek to synthesize samples of unknown classes to help distinguish those of known classes. However, almost all existing approaches are limited to learned features, which may not be sufficient to capture the abnormality in testing samples of unknown classes.

In this work, we propose to utilize gradient-based representations for open-set recognition. As an extension to our previous work~\cite{lee2020gradients}, we further validate the concept of \textit{confounding labels} to generate gradient-based representations to differentiate the inputs that a model is familiar with from those that are considered unknown. Rather than relying solely on the learned features from a model, we utilize gradients to gain insights regarding the amount of adjustments to its parameters necessary to properly represent given inputs. We empirically show that the obtained representations can be utilized in an open-set recognition setting where no sample of unknown classes is available during training to capture the distinction between the \textit{knowns} and the \textit{unknowns}.

\section{Related Work}
\vspace{-2mm}


\noindent \textbf{Open-Set Recognition~} Open-set recognition (OSR) aims to detect the \textit{unknowns} as well as to correctly classify the \textit{knowns}. Scheirer~\etal~\cite{scheirer2012toward_osr} proposed a Support Vector Machine-based approach to add an extra hyperplane in parallel to the originally obtained hyperplanes for known classes to differentiate the unknown classes. Bendale and Boult~\cite{bendale2016towards_osdn} proposed to replace the Softmax layer with the OpenMax layer to obtain the class probabilities of the \textit{unknowns}. Ge~\etal~\cite{ge2017generative_openmax} and Neal~\etal~\cite{neal2018counterfactual} proposed to utilize generative networks to create synthetic samples of the \textit{unknowns} and train open-set classifiers with them as an additional class. Yoshihashi~\etal~\cite{yoshihashi2019crosr} utilized latent representations obtained from an hierarchical reconstruction network for robust unknown detection. Oza and Patel~\cite{oza2019c2ae} employed class-conditioned auto-encoders with a novel training and testing setup. Sun~\etal~\cite{sun2020cgdl} proposed to learn conditional Gaussian distributions for known classification and unknown detection. However, almost all existing approaches rely on features learned from the \textit{knowns} to characterize the \textit{unknowns}.\vspace{1mm}

\noindent \textbf{Gradients~} Gradient-based optimization techniques~\cite{ruder2016gradient} have been at the core of numerous large-scale machine learning applications. Apart from their original utility as a tool to search for a converged solution, gradients have been utilized for various purposes, including visualization~\cite{zeiler2014visualizing, selvaraju2017gradcam, prabhushankar2020contrastive} and adversarial attack generation~\cite{goodfellow2014adversarial, madry2017pgd}. Gradients have also been explored to obtain effective representations~\cite{oberdiek2018classification, Kwon2019distorted, sun2020implicit, kwon2020backpropagated, lee2020gradients} for many applications including image quality and saliency estimation, and out-of-distribution/anomaly/novelty detection. However, the effectiveness of gradient-based representations has not been fully explored in the application of open-set recognition. 
\vspace{-2mm}
\section{Open-Set Recognition with Gradient-Based Representations}
\vspace{-2mm}

In this section, we introduce our framework for open-set recognition with gradient-based representations. We explain the setups to train and test relevant classifiers and unknown detectors, and we validate the effectiveness of gradient-based representations in open-set recognition settings. \vspace{-2mm}

\vspace{-2mm}
\subsection{Proposed Open-Set Recognition Framework}\label{ssec:osr_framework}
\vspace{-1mm}

In our previous work~\cite{lee2020gradients}, we introduced the framework to obtain gradient-based representations with \textit{confounding labels} for anomalous sample detection. We defined a \textit{confounding label} as a label that is different from ordinary labels on which a model is trained. Our intuition was that gradients correspond to the amount of change a model requires to properly represent a given sample. By introducing an unseen class label to the model with pre-defined representation space, the required model updates captured in gradients would be pertinent to mapping its relevant features to the new class if the model is familiar with the given input. If the input is not within the scope of the model, however, updates will be necessary for feature extraction and mapping, leading to a larger total amount of updates. 

Open-set recognition is a natural extension to our previous work by utilizing the gradient representations obtained from a classifier to train an unknown detector to reject the \textit{unknowns} while classifying the \textit{knowns} as the classifier was originally intended. We introduce our open-set recognition framework with gradient-based representations in Fig.~\ref{fig:framework}. Given a trained classification network $f(\theta)$ and an input image $x$, the network produces an output $f(x; \theta)$. Binary cross entropy loss is computed between the model output and a \textit{confounding label} $y_c$, which is a vector of length $N$ with $n$ number of $1$'s where $N$ is the number of classes in training and $n \in \{0, \dots, N\} \setminus \{1\}$. The loss is backpropagated to generate gradients $\nabla J(\theta; x, y_c)$, and gradient-based representation is formed by concatenating the magnitudes of gradients from every parameter set in the model as the following: 
\begin{equation}
  \begin{gathered}
    \big[~\|\nabla J_{\theta_0}(\theta; x, y_c)\|^2_2~,~\cdots~,~\|\nabla J_{\theta_{P-1}}(\theta; x, y_c)\|^2_2~\big], \vspace{2mm}\\
    \text{$P$: the number of parameter sets in a given network.}
  \end{gathered}
\end{equation}
Based on the decision of the unknown detector $d$, the final classification is determined: if known, the prediction of the original classifier is preserved ($c \in \{0, N - 1\}$); if unknown, then the model prediction is $N$. Overall, there are $N + 1$ possible options for final prediction.

\vspace{-3mm}
\subsection{Closed-Set Training \& Open-Set Testing}
\vspace{-1mm}

\begin{figure}[b]
    \centering
    \includegraphics[width=0.99\linewidth]{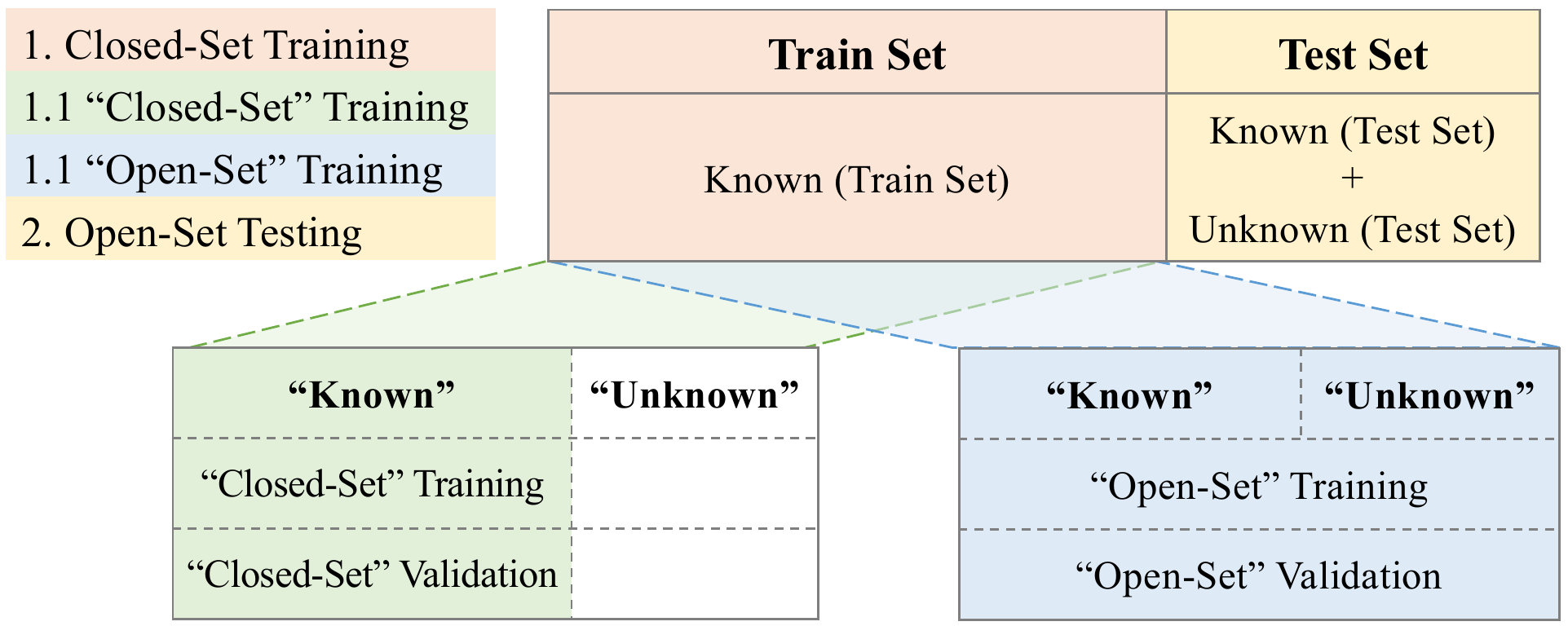}
    \caption{Data split for training and testing closed-set and open-set classifiers and unknown detectors.}
    \label{fig:data_split}
\end{figure}

The proposed method for open-set recognition has two stages: closed-set training and open-set testing. Closed-set includes the \textit{knowns}, while open-set includes both the \textit{knowns} and the \textit{unknowns}. The closed-set training phase is then split into two stages: ``closed-set'' training and ``open-set'' training, where we randomly select some of the known classes to be the ``unknowns'' for training. This data split protocol is described in Fig.~\ref{fig:data_split}, and we explain the details of each stage in this section with color coordination for clarity. \vspace{2mm}

\definecolor{apricot}{RGB}{250, 225, 209}
\definecolor{lightgreen}{RGB}{222, 238, 213}
\definecolor{lightblue}{RGB}{217, 232, 245}
\definecolor{lightyellow}{RGB}{255, 241, 199}

\noindent \textbf{Closed-Set Training~} In training any network for open-set recognition, whether it is a classifier or an unknown detector, we have no access to any samples of the \textit{unknowns}, $\mathcal{U}$, as opposed to the \textit{knowns}, $\mathcal{K}$. Therefore, the training samples of the \textit{knowns}, $\mathcal{K}_{train}$, are split into two groups: ``knowns'', $\mathcal{K_K}$, and ``unknowns'', $\mathcal{K_U}$. First, the \colorbox{lightgreen}{``known'' samples} $\in \mathcal{K_K}$ are used to train a ``closed-set'' classifier. Then gradient-based representations for \colorbox{lightblue}{all samples of the ``knowns'' and} \colorbox{lightblue}{``unknowns''} $\in \mathcal{K_K}~\cup~\mathcal{K_U} = \mathcal{K}_{train}$ can be collected to train and validate an ``open-set'' unknown detector. In addition, we train a closed-set classifier with the \colorbox{apricot}{training samples of all} \colorbox{apricot}{\textit{knowns}} $\in \mathcal{K}_{train}$ to be utilized in the open-set testing stage. \vspace{-2mm}

\noindent \textbf{Open-Set Testing~} For the testing of the open-set recognition framework, we now utilize the closed-set classifier and the ``open-set'' unknown detector, trained in the previous stage. First, we input the \colorbox{lightyellow}{test samples of both the \textit{knowns} and the} \colorbox{lightyellow}{\textit{unknowns}} $\in \mathcal{K}_{test} \cup \mathcal{U}_{test}$ into the closed-set classifier to collect model predictions as well as gradient-based representations, as described in Sec.~\ref{ssec:osr_framework}. The gradient-based representations are then passed to the trained ``open-set'' unknown detector to determine whether the samples may be among the \textit{knowns} or the \textit{unknowns}. Then, based on the detector prediction, the classification from the closed-set classifier for each test sample is preserved or replaced with a new class label representing the \textit{unknowns}.

\vspace{-3mm}
\subsection{Effectiveness of Gradient-based Representations}\label{ssec:gradients}
\vspace{-1mm}

\begin{figure*}[b]
\vspace{-3mm}
    \centering
    \begin{subfigure}[t]{.28\textwidth}
        \centering
        \includegraphics[height=5cm, trim=8mm 0 0 0, clip]{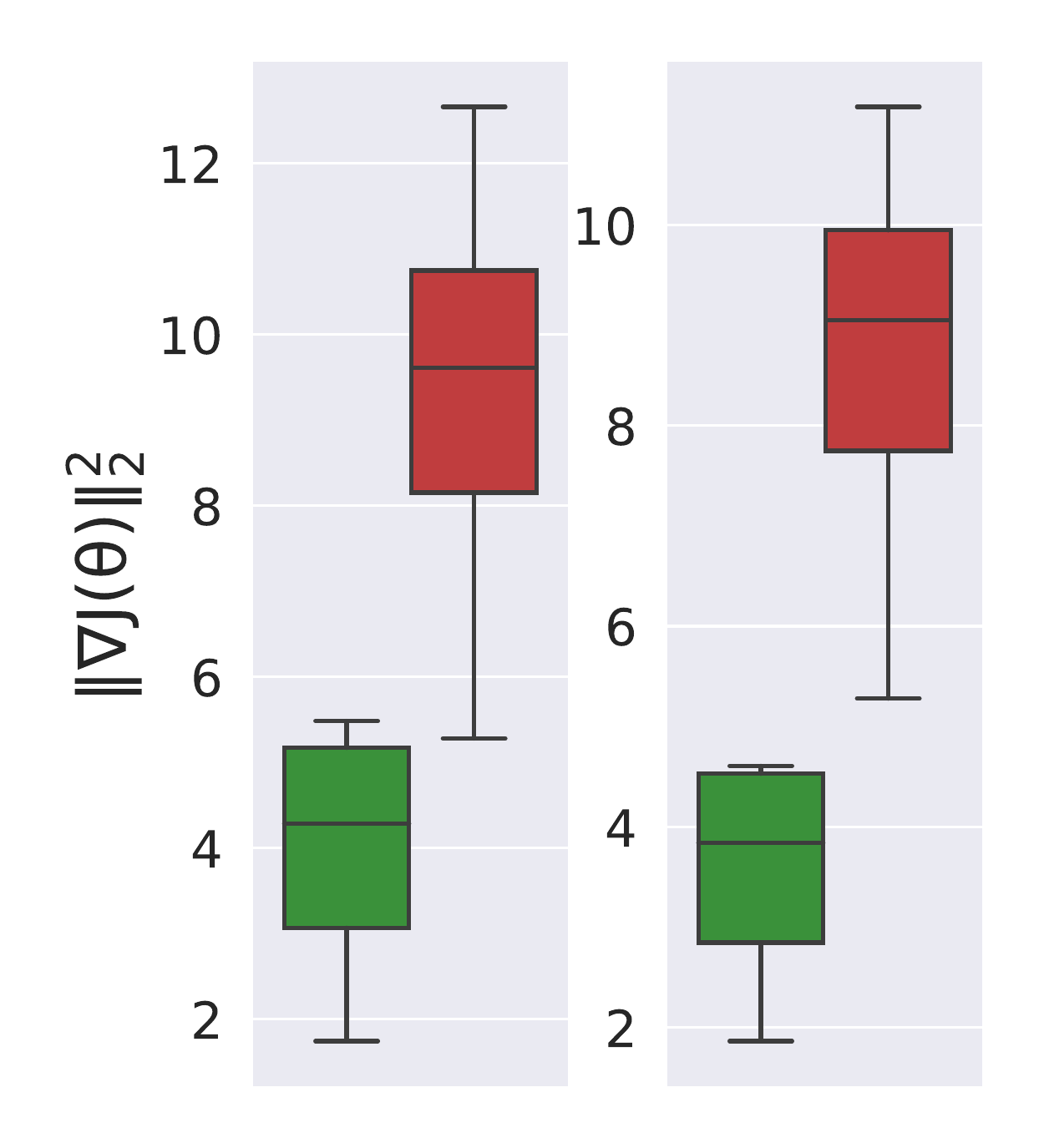}\vspace{-2mm}\\
        \hspace{5mm}\includegraphics[height=1.1cm]{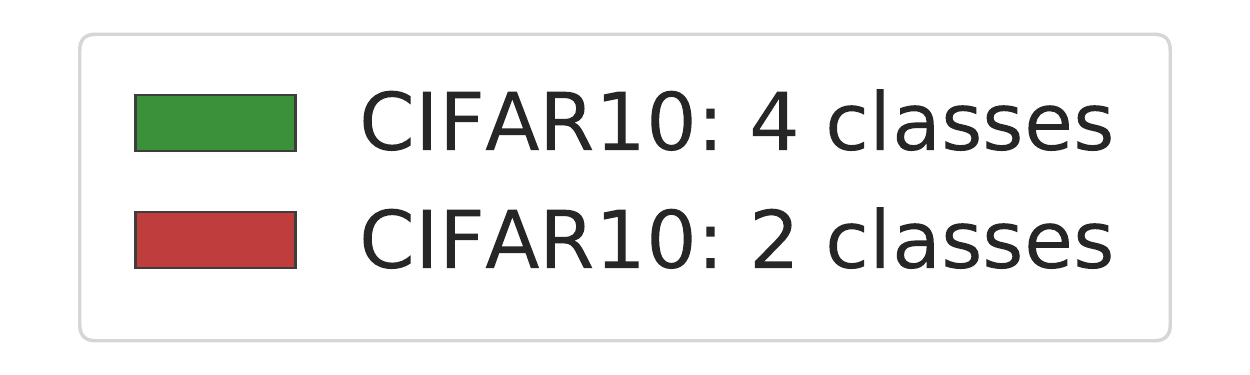}
        \subcaption{(CIFAR-10) 4 classes / 2 classes}
        \label{sfig:gradE_1}
    \end{subfigure}
    \begin{subfigure}[t]{.26\textwidth}
        \centering
        \includegraphics[height=5cm, trim=1.9cm 0 0 0, clip]{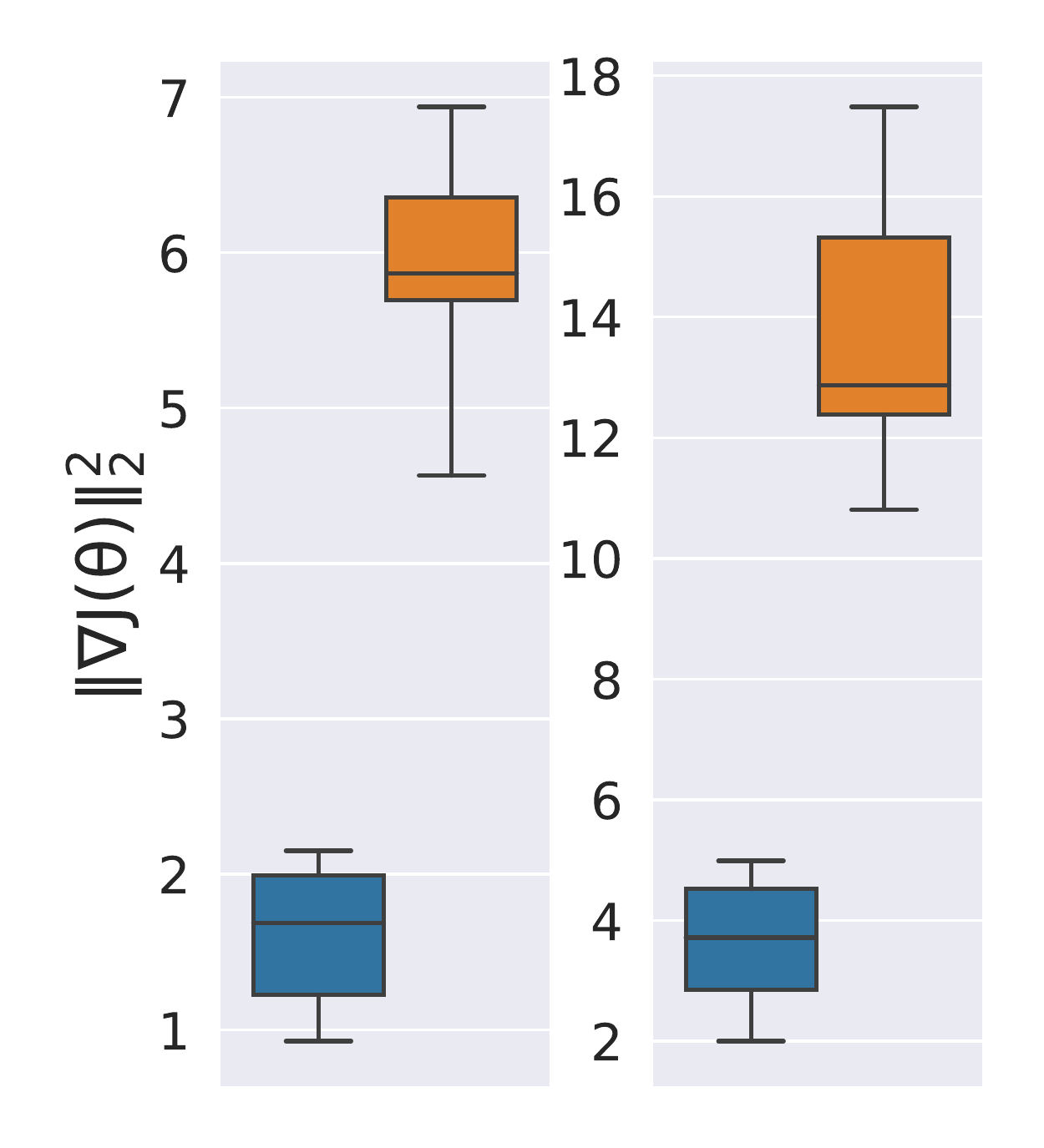}\vspace{-2mm}\\
        \includegraphics[height=1.1cm]{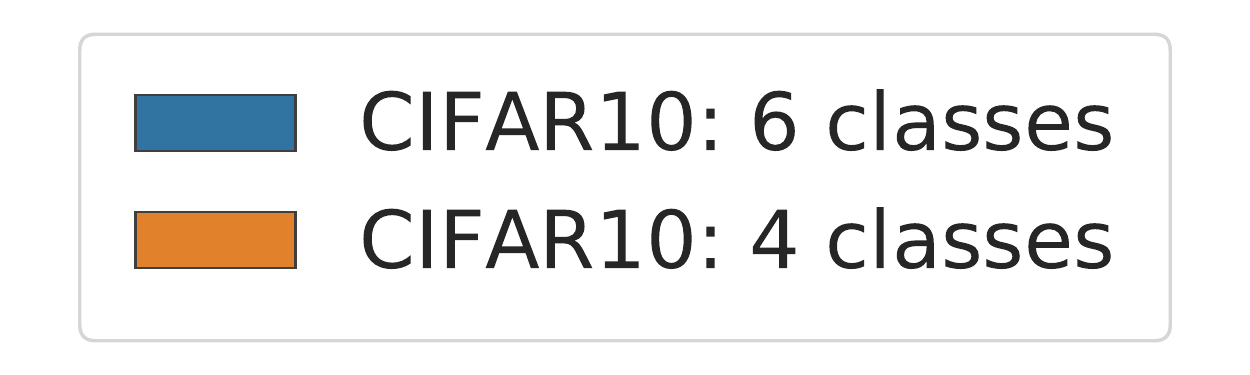}
        \subcaption{(CIFAR-10) 6 classes / 4 classes}
        \label{sfig:gradE_2}
    \end{subfigure}
    \begin{subfigure}[t]{.39\textwidth}
        \centering
        \includegraphics[height=5cm, trim=1.9cm 0 0 0, clip]{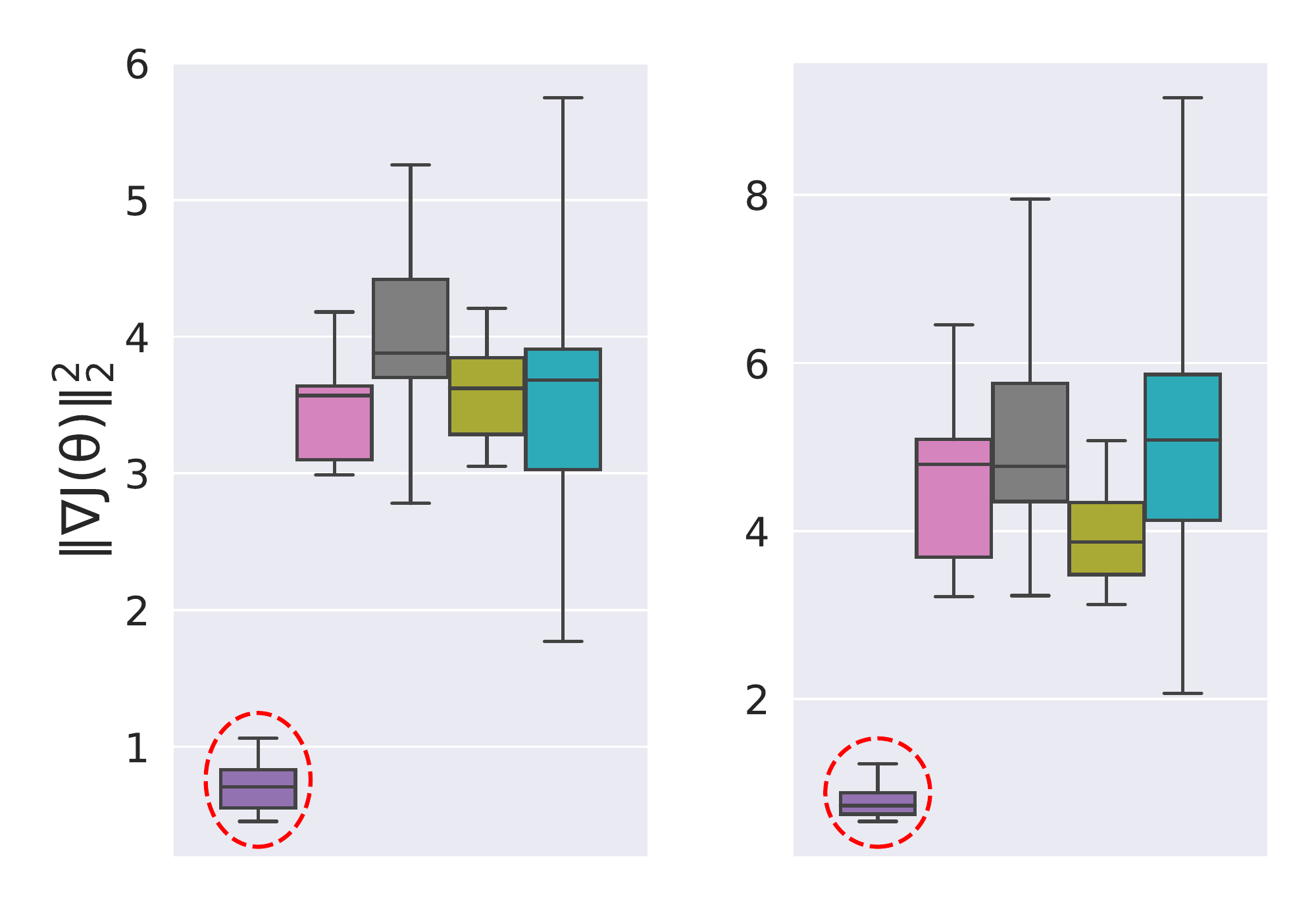} \vspace{-2mm}\\
        \includegraphics[height=1.1cm, trim=5mm 0 0 0, clip]{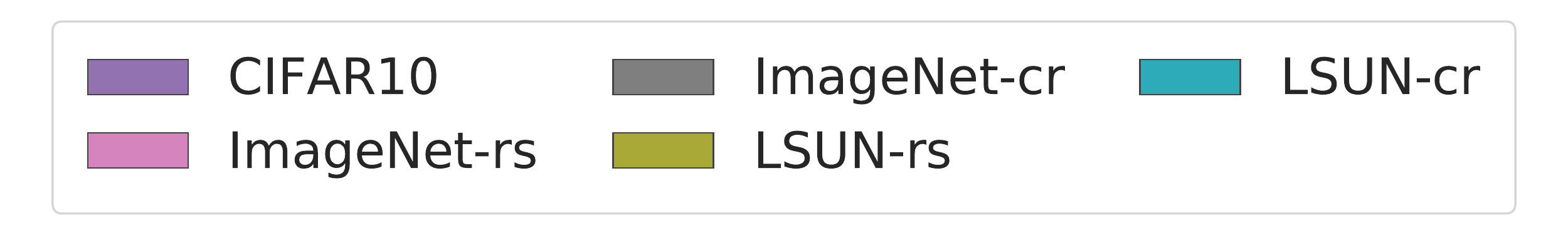}
        \subcaption{CIFAR-10 / ImageNet, LSUN}
        \label{sfig:gradE_3}
    \end{subfigure} \vspace{-2mm}
\caption{Comparison between the distributions of gradient magnitudes with \textit{knowns} / \textit{unknowns}. (cr = crop; rs = resize)}
\vspace{-2mm}
\label{fig:gradE}
\end{figure*}

 
In this section, we demonstrate the effectiveness of gradients in open-set recognition setting. We create mainly two testing scenarios: 1) the \textit{unknowns} chosen from the same dataset as the \textit{knowns}; 2) utilizing additional datasets as \textit{unknowns}. For the first scenario, we employ the random class splits of 6 known classes and 4 unknown classes on CIFAR-10~\cite{krizhevsky2009cifar} dataset, widely used for the evaluation of open-set recognition approaches. Among the first 6 classes, we also split them into 4 known classes and 2 unknown classes for further analysis. For the second scenario, we use ImageNet and LSUN (resize and crop) datasets as \textit{unknowns}, collected by~\cite{liang2017odin} and also conventionally used for open-set recognition. In each case, a ResNet-18 classifier is trained with the training set of the \textit{knowns} and the distributions of gradient magnitudes are collected on the test sets of the \textit{knowns} and the \textit{unknowns} to be visualized for 2 different model parameter sets in Fig.~\ref{fig:gradE}. The \textit{knowns} and the \textit{unknowns} for each figure are specified in the corresponding caption. It is clear that the gradient magnitudes for the \textit{knowns} are smaller than the \textit{unknowns} in every case. Comparing Fig.~\ref{sfig:gradE_1} and \subref{sfig:gradE_2}, however, there is a more clear distinction in gradient magnitudes between the \textit{knowns} and the \textit{unknowns} when the number of known classes and samples are larger. When the \textit{unknowns} are drawn from different datasets, as shown in Fig.~\ref{sfig:gradE_3}, the distributions of the \textit{knowns} and the \textit{unknowns} are clearly separated. The gradient magnitude distributions of the \textit{knowns} in Fig.~\ref{sfig:gradE_3} are highlighted in red circles for clarity. In all scenarios, gradients prove to be an effective tool to distinguish the \textit{unknowns} from the \textit{knowns}.

\vspace{-3.5mm}
\section{Experiments}
\vspace{-2mm}

In this section, we utilize the gradient-based representations obtained with \textit{confounding labels} for open-set recognition. The performance of open-set recognition methods is evaluated mainly in two aspects: open-set identification and open-set classification. Open-set identification focuses on the detection of the \textit{unknowns} when utilizing a single dataset to define the \textit{knowns} and the \textit{unknowns}. On the other hand, open-set classification focuses on the classification accuracy where the model makes predictions of $N + 1$ possible options, including the \textit{unknown} drawn from different datasets than the \textit{knowns}. The experiment setups expand upon the described scenarios in Sec.~\ref{ssec:gradients}, with Fig.~\ref{sfig:gradE_1} and \subref{sfig:gradE_2} concerned with open-set identification within a single dataset, while Fig.~\ref{sfig:gradE_2} and \subref{sfig:gradE_3} with open-set classification. We describe the details of each setup in the corresponding sections. For implementations, ResNet-18 with no pre-training is utilized as a classifier, and a binary classifier of 2 fully-connected layers is used as an unknown detector.





\subsection{Open-Set Identification}
\vspace{-1mm}

For open-set identification, we employ the widely accepted setup of selecting some classes at random to be used as \textit{knowns} and the remainder as \textit{unknowns}. In this scenario, the samples of the \textit{unknowns} come from the same dataset as the \textit{knowns}. As described in Sec.~\ref{ssec:gradients} regarding Fig.~\ref{sfig:gradE_1} and~\subref{sfig:gradE_2}, we first select 6 known classes $\in \mathcal{K}$ and 4 unknown classes $\in \mathcal{U}$ from the 10 overall classes of CIFAR-10. Then from $\mathcal{K}_{train}$, we select 4 ``known'' classes $\in \mathcal{K_K}$ and 2 ``unknown'' classes $\in \mathcal{K_U}$. With $\mathcal{K_K}$ and $\mathcal{K_U}$, we train an unknown detector and test it on $\in \mathcal{K}_{test} \cup \mathcal{U}_{test}$. We repeat for 5 different randomized sets of class splits, and we report the performance of unknown detectors in Table~\ref{tab:osi}. While our approach outperforms the existing methods prior to 2019 by a large margin, the more recent methods outperform our method. This is due to the performance of the unknown detectors trained for the 4 ``known'' and 2 ``unknown'' classes. As shown in Fig.~\ref{sfig:gradE_1} , the distinction between the gradient magnitude distributions is less significant when there are fewer number of ``known'' and ``unknown'' classes. The trained unknown detectors with the mentioned 4--2 class split show below 90\% accuracy on their validation set, leading to even lower discriminative results when evaluated with the unknown classes. 

\begin{table}[t]
\captionsetup{width=.95\linewidth}
\caption{Open-set identification results on CIFAR-10 dataset in AUROC. For methods other than the proposed method, we report the experimental results from~\cite{yoshihashi2019crosr,sun2020cgdl}.}
\centering
\resizebox{0.5\linewidth}{!}{%
\renewcommand{\arraystretch}{1.15}
\begin{tabular}{lc}
\hline \hline
 & \multicolumn{1}{c}{CIFAR-10} \\ \hline
Softmax & 0.677 \\
OpenMax~\cite{bendale2016towards_osdn} & 0.695 \\
G-OpenMax~\cite{ge2017generative_openmax} & 0.675 \\
OSRCI~\cite{neal2018counterfactual} & 0.699 \\
C2AE~\cite{oza2019c2ae} & 0.895 \\
CGDL~\cite{sun2020cgdl} & \textbf{0.903}\\ \hdashline
Ours & 0.838 \\ \hline \hline \vspace{-9mm}
\end{tabular} 
}\label{tab:osi}
\end{table}

\vspace{-3mm}
\subsection{Open-Set Classification}

\begin{table*}[b] \vspace{-1mm}
\captionsetup{width=.9\linewidth}
\caption{Open-set classification results on CIFAR-10 dataset with various outliers added to the test set as \textit{unknowns}. For methods other than the proposed method, we report the experimental results from~\cite{yoshihashi2019crosr,sun2020cgdl}.}
\centering
\resizebox{0.7\textwidth}{!}{%
\renewcommand{\arraystretch}{1.15}
\begin{tabular}{lcccc}
\hline \hline 
 & \multicolumn{1}{c}{ImageNet-resize} & \multicolumn{1}{c}{ImageNet-crop} & \multicolumn{1}{c}{LSUN-resize} & \multicolumn{1}{c}{LSUN-crop} \\ \hline
Softmax & 0.653 & 0.639 & 0.647 & 0.642 \\
OpenMax~\cite{bendale2016towards_osdn} & 0.684 & 0.660 & 0.668 & 0.657 \\
LadderNet+OpenMax~\cite{yoshihashi2019crosr} & 0.670 & 0.653 & 0.659 & 0.652 \\
DHRNet+OpenMax~\cite{yoshihashi2019crosr} & 0.675 & 0.655 & 0.664 & 0.656 \\
CROSR~\cite{yoshihashi2019crosr} & 0.735 & 0.721 & 0.749 & 0.720 \\
C2AE~\cite{oza2019c2ae} & 0.826 & 0.837 & 0.801 & 0.783 \\ 
CGDL~\cite{sun2020cgdl} & 0.832 & 0.840 & 0.812 & 0.806 \\ \hdashline
Ours & \textbf{0.842} & \textbf{0.912} & \textbf{0.882} & \textbf{0.922} \\ \hline \hline
\end{tabular} \vspace{-8mm}
}\label{tab:osc}
\end{table*}

For open-set classification, we utilize CIFAR-10 dataset as the \textit{knowns} and ImageNet and LSUN (resize and crop) as the \textit{unknowns}. Specifically, the test set of each dataset being used as the \textit{unknowns} is added to the test set of CIFAR-10. Each dataset as well as the test set of CIFAR-10 contains 10,000 testing samples, making the known-to-unknown ratio 1:1. As described in Sec.~\ref{ssec:gradients} regarding Fig.~\ref{sfig:gradE_2} and~\subref{sfig:gradE_3}, we create 6--4 class split using the train set of CIFAR-10 to train an unknown detector. We repeat for 5 different randomized sets of class splits, and we report the open-set classification accuracy on the combined test sets of the \textit{knowns} and the \textit{unknowns} in Table~\ref{tab:osc}. During the open-set identification experiments in the previous section, we noticed that the thresholding values for the output of binary classifiers to achieve the best AUROC scores are higher than the regular 0.5, averaging over 0.9, when the detectors trained on 4--2 class split are evaluated on the combined 6--4 class split. Based on this observation, we fix the thresholding value for unknown detectors to 0.95 during open-set classification testing. Our approach with gradient-based representations outperforms all recent methods by a large margin. The unknown detector shows better performance when the \textit{unknowns} come from different datasets than the classifier training dataset, similar to the out-of-distribution detection setup in our previous work~\cite{lee2020gradients}. The exceptional out-of-distribution detection performance with the gradient-based representations is matched with the performance of the unknown detectors in this work. These results prove that gradients prove to be an effective tool to capture the \textit{unknowns} from the \textit{knowns} in open-set recognition setting.

\vspace{-2mm}
\section{Conclusion}
\vspace{-1mm}

In this paper, we utilized gradient-based representations obtained from a trained classifier with \textit{confounding labels} to detect samples of the \textit{unknowns} while preserving model predictions for those detected to be the \textit{knowns}. We empirically show that the obtained representations can be utilized in an open-set recognition setting where no sample of the \textit{unknowns} is available during training to capture the distinction between the \textit{knowns} and the \textit{unknowns} by exploiting the training samples of the \textit{knowns}. We validate our approach on open-set identification and classification.
\newpage

\small
\bibliographystyle{IEEEbib}
\bibliography{bib}

\begin{thebibliography}{10}

\bibitem{temel2018cureor}
Dogancan Temel, Jinsol Lee, and Ghassan AlRegib,
\newblock ``Cure-or: Challenging unreal and real environments for object
  recognition,''
\newblock in {\em 2018 17th IEEE International Conference on Machine Learning
  and Applications (ICMLA)}. IEEE, 2018, pp. 137--144.

\bibitem{temel2019multifarious}
Dogancan Temel, Jinsol Lee, and Ghassan AlRegib,
\newblock ``Object recognition under multifarious conditions: A reliability
  analysis and a feature similarity-based performance estimation,''
\newblock in {\em 2019 IEEE International Conference on Image Processing
  (ICIP)}. IEEE, 2019, pp. 3033--3037.

\bibitem{goodfellow2014adversarial}
Ian~J Goodfellow, Jonathon Shlens, and Christian Szegedy,
\newblock ``Explaining and harnessing adversarial examples,''
\newblock {\em arXiv preprint arXiv:1412.6572}, 2014.

\bibitem{guo2017calibration}
Chuan Guo, Geoff Pleiss, Yu~Sun, and Kilian~Q Weinberger,
\newblock ``On calibration of modern neural networks,''
\newblock in {\em Proceedings of the 34th International Conference on Machine
  Learning-Volume 70}. JMLR. org, 2017, pp. 1321--1330.

\bibitem{scheirer2012toward_osr}
Walter~J Scheirer, Anderson de~Rezende~Rocha, Archana Sapkota, and Terrance~E
  Boult,
\newblock ``Toward open set recognition,''
\newblock {\em IEEE transactions on pattern analysis and machine intelligence},
  vol. 35, no. 7, pp. 1757--1772, 2012.

\bibitem{bendale2016towards_osdn}
Abhijit Bendale and Terrance~E Boult,
\newblock ``Towards open set deep networks,''
\newblock in {\em Proceedings of the IEEE conference on computer vision and
  pattern recognition}, 2016, pp. 1563--1572.

\bibitem{yoshihashi2019crosr}
Ryota Yoshihashi, Wen Shao, Rei Kawakami, Shaodi You, Makoto Iida, and Takeshi
  Naemura,
\newblock ``Classification-reconstruction learning for open-set recognition,''
\newblock in {\em Proceedings of the IEEE Conference on Computer Vision and
  Pattern Recognition}, 2019, pp. 4016--4025.

\bibitem{oza2019c2ae}
Poojan Oza and Vishal~M Patel,
\newblock ``C2ae: Class conditioned auto-encoder for open-set recognition,''
\newblock in {\em Proceedings of the IEEE Conference on Computer Vision and
  Pattern Recognition}, 2019, pp. 2307--2316.

\bibitem{sun2020cgdl}
Xin Sun, Zhenning Yang, Chi Zhang, Keck-Voon Ling, and Guohao Peng,
\newblock ``Conditional gaussian distribution learning for open set
  recognition,''
\newblock in {\em Proceedings of the IEEE/CVF Conference on Computer Vision and
  Pattern Recognition}, 2020, pp. 13480--13489.

\bibitem{ge2017generative_openmax}
ZongYuan Ge, Sergey Demyanov, Zetao Chen, and Rahil Garnavi,
\newblock ``Generative openmax for multi-class open set classification,''
\newblock {\em Proceedings of the British Machine Vision Conference (BMVC)},
  pp. 42.1--42.12., 2017.

\bibitem{neal2018counterfactual}
Lawrence Neal, Matthew Olson, Xiaoli Fern, Weng-Keen Wong, and Fuxin Li,
\newblock ``Open set learning with counterfactual images,''
\newblock in {\em Proceedings of the European Conference on Computer Vision
  (ECCV)}, 2018, pp. 613--628.

\bibitem{lee2020gradients}
Jinsol Lee and Ghassan AlRegib,
\newblock ``Gradients as a measure of uncertainty in neural networks,''
\newblock in {\em 2020 IEEE International Conference on Image Processing
  (ICIP)}. IEEE, 2020, pp. 2416--2420.

\bibitem{ruder2016gradient}
Sebastian Ruder,
\newblock ``An overview of gradient descent optimization algorithms,''
\newblock {\em arXiv preprint arXiv:1609.04747}, 2016.

\bibitem{zeiler2014visualizing}
Matthew~D Zeiler and Rob Fergus,
\newblock ``Visualizing and understanding convolutional networks,''
\newblock in {\em European conference on computer vision}. Springer, 2014, pp.
  818--833.

\bibitem{selvaraju2017gradcam}
Ramprasaath~R Selvaraju, Michael Cogswell, Abhishek Das, Ramakrishna Vedantam,
  Devi Parikh, and Dhruv Batra,
\newblock ``Grad-cam: Visual explanations from deep networks via gradient-based
  localization,''
\newblock in {\em Proceedings of the IEEE international conference on computer
  vision}, 2017, pp. 618--626.

\bibitem{prabhushankar2020contrastive}
Mohit Prabhushankar, Gukyeong Kwon, Dogancan Temel, and Ghassan AlRegib,
\newblock ``Contrastive explanations in neural networks,''
\newblock in {\em 2020 IEEE International Conference on Image Processing
  (ICIP)}. IEEE, 2020, pp. 3289--3293.

\bibitem{madry2017pgd}
Aleksander Madry, Aleksandar Makelov, Ludwig Schmidt, Dimitris Tsipras, and
  Adrian Vladu,
\newblock ``Towards deep learning models resistant to adversarial attacks,''
\newblock {\em arXiv preprint arXiv:1706.06083}, 2017.

\bibitem{oberdiek2018classification}
Philipp Oberdiek, Matthias Rottmann, and Hanno Gottschalk,
\newblock ``Classification uncertainty of deep neural networks based on
  gradient information,''
\newblock in {\em IAPR Workshop on Artificial Neural Networks in Pattern
  Recognition}. Springer, 2018, pp. 113--125.

\bibitem{Kwon2019distorted}
Gukyeong Kwon, Mohit Prabhushankar, Dogancan Temel, and Ghassan AlRegib,
\newblock ``Distorted representation space characterization through
  backpropagated gradients,''
\newblock in {\em 2019 26th IEEE International Conference on Image Processing
  (ICIP)}, 2019.

\bibitem{sun2020implicit}
Yutong Sun, Mohit Prabhushankar, and Ghassan AlRegib,
\newblock ``Implicit saliency in deep neural networks,''
\newblock in {\em 2020 IEEE International Conference on Image Processing
  (ICIP)}. IEEE, 2020, pp. 2915--2919.

\bibitem{kwon2020backpropagated}
Gukyeong Kwon, Mohit Prabhushankar, Dogancan Temel, and Ghassan AlRegib,
\newblock ``Backpropagated gradient representations for anomaly detection,''
\newblock {\em arXiv preprint arXiv:2007.09507}, 2020.

\bibitem{krizhevsky2009cifar}
Alex Krizhevsky,
\newblock ``Learning multiple layers of features from tiny images,''
\newblock Tech. {R}ep., 2009.

\bibitem{liang2017odin}
Shiyu Liang, Yixuan Li, and Rayadurgam Srikant,
\newblock ``Enhancing the reliability of out-of-distribution image detection in
  neural networks,''
\newblock {\em arXiv preprint arXiv:1706.02690}, 2017.

\end{thebibliography}

\end{document}